\documentclass{article}

\usepackage{booktabs}

\usepackage{ijcai18}

\usepackage{times}
\usepackage{xcolor}
\usepackage{soul}
\usepackage[utf8]{inputenc}
\usepackage[small]{caption}

\usepackage{paralist,subfig}
\usepackage{enumitem}
\usepackage{amsmath, amssymb, amsmath, mathtools, amsthm}%amsthm
\usepackage{graphicx, comment, epstopdf, multirow}
\usepackage{algorithm, algorithmicx}
\usepackage{algpseudocode} 
\usepackage{tikz, pgfplots, epstopdf, todonotes,circuitikz}
\usetikzlibrary{positioning}
\usepackage{colortbl}
\definecolor{Gray}{gray}{0.90}

\DeclareMathOperator*{\argmin}{argmin}
\DeclareMathOperator*{\argmax}{argmax}

\newcommand{\norm}[1]{\left\lVert#1\right\rVert}

\newcommand{\ty}[0]{\tilde{y}}
\newcommand{\taus}[0]{\tau_s}
\newcommand{\fs}[0]{f_s}

\newcommand{\s}[0]{^{(s)}}

\algnewcommand\False{\textbf{false}\space}
\algnewcommand\True{\textbf{true}\space}

\algnewcommand{\algorithmicgoto}{\textbf{go to}}%
\algnewcommand{\Goto}[1]{\algorithmicgoto~#1}%

\newtheorem{deff}{Definition}
\newtheorem{propp}{Proposition}
\begin{document}

%\title{Evaluating Stealthy Attacks on Regression-Based Anomaly Detectors Used in CPS}  % put your title here!
\title{Adversarial Regression for Detecting Attacks in~Cyber-Physical~Systems}  % put
                                % your title here!

\author{Amin Ghafouri \and Yevgeniy Vorobeychik \and Xenofon
  Koutsoukos\\Electrical Engineering and Computer Science\\Vanderbilt
  University\\Nashville, TN\\\{amin.ghafouri,yevgeniy.vorobeychik,xenofon.koutsoukos\}@vanderbilt.edu}

%\keywords{anomaly detection; stealthy attack; machine learning regression; process control system}  % put your semicolon-separated keywords here!

\maketitle

\begin{abstract} 
Attacks in cyber-physical systems (CPS) which manipulate sensor
readings can cause enormous physical damage if undetected.
Detection of attacks on sensors 
is crucial to mitigate this issue.
We study supervised regression as a means to detect anomalous sensor
readings, where each sensor's measurement is predicted as a function
of other sensors.
We show that several common learning approaches in this context
are still vulnerable to \emph{stealthy attacks}, which carefully
modify readings of compromised sensors to cause desired damage while
remaining undetected.
Next, we model the interaction between the CPS defender and attacker
as a Stackelberg game in which the defender chooses detection
thresholds, while the attacker deploys a stealthy attack in response.
We present a heuristic algorithm for finding an approximately optimal threshold for
the defender in this game, and show that it increases system
resilience to attacks without significantly increasing the false alarm rate.
\end{abstract}

%%%%%%%%%%%%%%%%%%%%%%%%%%%%%%%%%%%%%%%%%%%%%%%%%%%%%%%%%%%%%%%%%%%%%%%%%%%%%%%%%%%%%%%%%%%%%%%%%%%%%%%%%
%% start of main body of paper

\section{Introduction}\label{ch:optimal}

Cyber-physical systems (CPS) form the foundation of much of the critical infrastructure, including the electric power grid, transportation networks, water networks, and nuclear plants, among others.
Malicious attacks on CPS can consequently lead to major disruptions, from major blackouts to nuclear incidents.
Moreover, such attacks are no longer hypothetical, with Stuxnet perhaps the best-known example of an attack targeting physical infrastructure through cyber means.
A crucial aspect of the Stuxnet attack---and a feature that is central to most attacks on CPS---is the corruption of sensor readings to either ensure that an attack on the controller code is undetected, or indirectly impact controller behavior.
The power of sensor corruption to impact control decisions is particularly alarming: since controllers are commonly making decisions directly as a consequence of readings from \emph{critical sensors}, for example, increasing or decreasing temperature to ensure it stays within a target (safe) range, tampering with these readings (e.g., temperature) can cause the controller itself to put the system into an unsafe state.
With no remediation, as soon as a critical sensor is compromised, the attacker can cause major system failure by directly manipulating sensor measurements in essentially arbitrary ways.

Data-driven anomaly detection techniques have previously been proposed to mitigate against arbitrary manipulations of sensor data, both based on temporal statistics of single-sensor measurements~\cite{ghafouri2016optimal}, as well as by modeling joint measurements to obtain a prediction for each sensor based on measurements of others~\cite{ghafouri2017optimal}.
However, these have been limited in two ways: 1) they generally consider a specific anomaly detection method, such as a Gaussian Process regression, and 2) they do not consider attacks which can arbitrarily modify a collection of sensor measurements.
%, specifically so as to deploy a successful attack while avoiding detection.

To address these limitations, we present a general framework for \emph{adversarial anomaly detection} in the context of integrity attacks on a subset of sensors in CPS.
We start with a general anomaly detection framework based on a collection of supervised regression models which predict a measurement for each sensor as a function of readings from other sensors.
We instantiate this framework with three regression models: linear regression, neural network, and an ensemble of the two.
We then develop an optimization approach for stealthy attacks on linear regression which aim to maximally increase or decrease a reading for a collection of target sensors (for example, those serving as direct inputs into a controller which maintains the system in a safe state), and extend these to consider a neural network detector, and the ensemble of these.

Next, we model robust anomaly detection as a Stackelberg game between the defender who chooses detection thresholds for each sensor to balance false alarms with impact from successful attacks, and the attacker who subsequently deploys a stealthy attack.
Finally, we develop a heuristic algorithm to compute a resilient detector in this model.
Our experiments show that (a) the stealthy attacks we develop are extremely effective,
%---what's particularly surprising is that they are more effective when predictions are based on a neural network than linear regression; 
and (b) our resilient detector significantly reduces the impact of a stealthy attack without appreciably increasing the false alarm rate.
\paragraph{Related Work}\label{sec:related}

A number of techniques have been proposed for anomaly detection in CPS, aimed at identifying either faults or attacks.
Several of these consider simplified mathematical models of the physical system, rather than using past data of normal behavior to identify anomalies~\cite{cardenas2011attacks,urbina2016limiting}.
A number of others use statistical and machine learning techniques for anomaly detection~\cite{nader2014l_p,junejo2016behaviour,ghafouri2016optimal,ghafouri2017optimal}, but restrict attention to a specific model, such as Gaussian Process regression~\cite{ghafouri2017optimal}, to a single sensor~\cite{ghafouri2016optimal}, to unsupervised techniques~\cite{nader2014l_p}, or to directly predict attacks based on normal and simulated attack data~\cite{junejo2016behaviour}.
Ours is the first approach which deals with a general framework for regression-based modeling of sensor readings based on other sensors.

In addition, considerable literature has developed around the general problem of adversarial machine learning, which considers both attacks on machine learning methods and making learning more robust~\cite{lowd2005adversarial,dalvi2004adversarial,Biggio13,li14,Vorobeychik14,Li18}.
However, all focus on inducing mistakes in a single model---almost universally, a classifier (\citeauthor{Grosshans13}~[\citeyear{Grosshans13}] is a rare exception).
In contrast, we study stealthy attacks on sensor readings, where even the critical measurement can be manipulated directly, but where multiple models used for anomaly detection impose strong constraints on which sensor manipulations are feasible.
As such, our framework is both conceptually and mathematically quite different from prior research on adversarial learning.  
\section{Regression-Based Anomaly Detection}\label{sec:model}

Consider a collection of $d$ sensors, with $y$ representing a vector of sensor measurements, where $y_s$ is the measurement of a sensor $s$.
In practice, such measurements often feed directly into controllers which govern the state of critical system components.
For example, such a controller may maintain a safe temperature level in a nuclear reactor by cooling the reactor when the temperature reading becomes too high, or heating it when it's too low.
If an attacker were to take control of the sensor which measures the temperature of the nuclear reactor, they can transmit low sensor readings to the controller, causing it to heat up the reactor to a temperature arbitrarily above a safe level!

A way to significantly reduce the degree of freedom available to the attacker in this scenario is to use anomaly detection to identify suspicious measurements.
To this end, we can leverage relationships among measurements from multiple sensors: if we can effectively predict a reading from a sensor based on the measurements of others, many sensors would need to be compromised for a successful attack, and even then the attacker would face complex constraints in avoiding detection: for example, slightly modifying the one sensor may actually cause \emph{another} sensor to appear anomalous.

\begin{table}[t!]
	\caption{List of Symbols}
	\label{tab:symbols}
	\centering
	\renewcommand*{\arraystretch}{1.0}
	\setlength{\tabcolsep}{3pt}
	\begin{tabular}{| c | p{6.2cm} |}
		\hline
		Symbol & \multicolumn{1}{l|}{Description} \\
		\hline
		\rowcolor{Gray} $S$ & Set of sensors \\
		$y_s$ & Actual measurement of sensor $s$ \\
		\rowcolor{Gray} $\ty_s$ & Observed measurement value of sensor $s$ \emph{after the attack} \\
%		$\hy_s$ & Predicted value of sensor $s$ \\ 
		$r_s$ & residual of sensor $s \in S$ \\
		\rowcolor{Gray}  $S_c$ & Set of critical sensors\\
		$\delta_s$ & Error added to sensor $s \in S$ \\
		\rowcolor{Gray} $B$ & The maximum number of sensors that can be attacked (attacker's budget)\\
		 $D$ & Set of sensors with detectors \\ 
		\rowcolor{Gray} $f_s$ & Regression-based predictor of detector $s \in D$ \\
		$\taus$ & Threshold value of detector $s \in D$ \\
		
		\hline
	\end{tabular}
\end{table}

We therefore propose the following general framework for anomaly detection in a multi-sensor CPS.
For each sensor $s$, we learn a model $f_s(y_{-s})$ which maps observed readings from sensors other than $s$ to a \emph{predicted} measurement of $s$.\footnote{Note that it is direct to extend such models to use other features, such as time and control outputs.}
%, $\hat{y}_s$
Learning can be accomplished by collecting a large amount of \emph{normal} system behavior for all sensors, and training the model for each sensor.
At operation time, we can then use these predictions, together with observed measurements, to check for each sensor $s$ whether its measurement is anomalous based on residual $r_s = |f_s(y_{-s}) - y_s|$, and a threshold $\tau_s$, where $r_s \le \tau_s$ is considered \emph{normal}, while $r_s > \tau_s$ is flagged as \emph{anomalous}.
For easy reference, we provide the list of these and other symbols used in the paper in Table~\ref{tab:symbols}.

In principle, any supervised learning approach can be used to learn the functions $f_s$.
We consider three illustrative examples: linear regression, neural network, and an ensemble of the two.

\paragraph{Linear Regression}

%Suppose that the predictor of each ML regression-based anomaly detector implements a linear regression model. In other words, 
In this case, for each detector $s$, we let the predictor be a linear regression model defined as $f_s(y_{-s}) = w_s^Ty_{-s} + b_s$, where $w_s$ and $b_s$ are the parameters of the linear regression model. 

\paragraph{Neural Network}

%Next, suppose the predictor of each ML regression-based detector implements a neural network regression model. 
In this case, for each sensor $s$, we let the predictor $\fs$ be a feed-forward neural network model defined by parameters $\theta_s$, where the prediction $f_s(y_{-s};\theta_s)$ is obtained by performing a forward-propagation. 

\paragraph{Ensemble}

It has been shown in the adversarial learning literature that multiple detectors can improve adversarial robustness~\cite{biggio2010multiple}.
We explore this idea for the ML regression-based detector by considering an ensemble model for the predictor. 
We consider an ensemble predictor that contains a neural network and a linear regression model. Different methods can be used for combining the results, but we consider a bagging approach where the result is computed as the average of the predictors' outputs. 
\section{Attacking the Anomaly Detector}\label{sec:advregression}

%\todo{in this section}
%In this section, we present the adversarial regression problem for
%CPS. The idea is to find a stealthy sensor attack that maximizes the
%deviation from the actual value for some critical sensor. We describe
%this in more detail below. 
%Figure~\ref{fig:diagram} illustrates the adversarial regression
%problem. 

We have informally argued above that the anomaly detector makes
successful sensor manipulation attacks challenging.
For example, the trivial attack on the critical sensors is now
substantially constrained.
We now study this issue systematically, and develop algorithmic
approaches for \emph{stealthy} attacks on sensors which attempt to
optimally skew measured values of target sensors without tripping an
anomaly detection alarm.
While arbitrary changes to the sensor readings are no longer feasible,
our experiments subsequently show that, unfortunately, a baseline
anomaly detector is still quite vulnerable.
Subsequently, we develop an approach for making it more robust to
stealthy attacks.

%To simplify the notation that is used in the next sections, we write $f\s(\ty_{-s})$ as $f\s(\ty)$ and assume that $f\s$ discards $\ty_s$ before prediction. %We prefer the latter approach as it makes the representations easier in the next sections.

\subsection{Attacker Model}
%\todo{attack model}
We define the attack model by describing the attacker's capability,
knowledge, and objective.
%, and strategy.
%~\cite{biggio2014security}.
\textbf{Capability:} The adversary may compromise and make arbitrary
modifications to the readings from up to $B$ sensors.
%The cardinality of this subset has an upper
%bound $B$. 
\textbf{Knowledge:} We consider a worst-case scenario where
the attacker has complete knowledge of the system and anomaly detectors. \textbf{Objective:} The attacker's
objective is to maximize or minimize the observed value for some
critical sensor $\bar{s}$ among a collection of target sensors $S_c$. 
%\textbf{Strategy:} The attack must remain undetected during its
%entire duration.
In addition, the attacker's modifications to the sensors it
compromises must remain undetected (stealthy) with respect to the anomaly
detection model.

%\todo{explaining the problem)
%Further, we let $S_c$ be a set of critical sensors, where a critical sensor is one whose compromise can affect the safety of the system (e.g., pressure of a reactor in process control systems). 

Formally, let $\ty$ be the sensor measurements \emph{observed after
  the attacker compromises a subset of sensors}.
%Formally, the attacker transmits corrupted measurements $\ty$ for the
%sensors it has compromised to the controller. 
We assume that the attacker can compromise at most $B$ sensors;
consequently, for the remaining $d-B$ sensors, $\ty_s = y_s$.
Additionally, the attacker cannot modify individual
measurements of compromised sensors arbitrarily (for example, constrained by
sensor-level anomaly detectors), so that $|\ty_s - y_s| \le \eta_s$;
$\eta_s = \infty$ is then a special case in which the attacker is not
so constrained.
The final set of constraints faced by the attacker ensures that the
attack is stealthy: that is, it doesn't trigger alerts based on the
anomaly detection model described in Section~\ref{sec:model}.
To accomplish this, the attacker must ensure that for any
sensor $s$, $| \ty_s - \fs(\ty_{-s}) | \leq \taus$, that is, the
corrupted measurements must \emph{simultaneously} appear normal with
respect to the detector model described above and observed
measurements $\ty_s$ for all sensors (both attacked, and not).
Note that this is a subtle constraint, since corrupted measurements
for one sensor may make \emph{another} appear anomalous.

For a given critical sensor $\bar{s}$, the attacker's goal is to maximize or minimize
the corrupted measurement $\ty_{\bar{s}}$.
Here we focus on minimization of observed values of critical sensors;
approaches where the goal is to maximize these values are essentially
the same.
%almost identical approaches work for minimization as well.
%The attacker's optimization problem in which he aims to maximize
%observed measurement of some critical sensor can then be described as
%follows:
The attacker's optimization problem is then
\begin{subequations}\label{eq:adv}
	\begin{align}
		\nonumber \argmin_{\bar{s} \in S_c} \; &  \min_{\tilde{y}} \tilde{y}_{\bar{s}} \\
		\textup{s.t.}\quad & | \tilde{y}_s - \fs(\ty_{-s}) | \leq \taus , \forall s \in D \label{c:stealth}\\
		& |\ty_s - y_s| \leq \eta_s, \forall{s} \label{c:bound}\\
		& \norm{\tilde{y} - y}_0  \leq B,\label{c:budget}
	\end{align}
\end{subequations}
where Constraints~\eqref{c:stealth} capture evasion of the anomaly
detector (to ensure that the attack is stealthy),
Constraints~\eqref{c:bound} limit modifications of individual sensors,
if relevant, and Constraint~\eqref{c:budget} limits the attacker to
attack at most $B$ sensors.

Not surprisingly, the problem faced by the adversary is NP-Hard.
\begin{propp}\label{prop:nphard}
The Adversarial Regression Problem is NP-Hard.
%\footnote{The proof is  provided in the extended version of the paper.}
\end{propp}
The proof of this proposition is in the appendix.
%% TODO: add link to extended version

Despite the hardness result, we now present algorithmic approaches for
solving Problem~\eqref{eq:adv} in the context of linear regression and
neural networks, as well as a simple ensemble of these described in
Section~\ref{sec:model}.

\subsection{Attacking Linear Regression}\label{sec:advlinear}

First, we replace the non-convex budget constraint using a collection of linear constraints.
Define $\alpha_s$ as a binary variable indicating whether
the sensor $s$ is attacked. 
Thus, Constraint~\eqref{c:budget} can
be re-written as $\sum_{i=1}^m \alpha_i \leq B$.  
Now, for each sensor $s$ let $\ty_s = y_s + \delta_s$, where $\delta_s$ 
represents the modification made by the adversary.
Since $\delta_s > 0$ only if $s$ has been compromised (i.e., $\alpha_s
= 1$), we add a constraint
%Further, we can remove $\alpha_i$ from each $\alpha_i \delta_i$ term,
%and instead add a new constraint 
$\delta_s \leq M \alpha_s$ where $M$ is a sufficiently large number.

Next, we rewrite Constraints~\eqref{c:stealth}.
In the case of a linear regression model for anomaly detection, 
%the
%first constraint of the adversarial regression problem~\eqref{eq:adv}
%becomes a set of linear equations. 
%To begin, 
Constraints~\eqref{c:stealth} become $| \tilde{y}_s - w_s^T
\tilde{y}_{-s} - b_s|  \leq \tau_s$.
These can be represented using two sets of linear constraints:
$\tilde{y}_s - w_s^T
\tilde{y}_{-s} - b_s  \leq \tau_s$ and $-\tilde{y}_s + w_s^T
\tilde{y}_{-s} + b_s  \leq \tau_s$.
Finally, Constraints~\eqref{c:bound} are similarly linearized as
$\delta_s \leq \eta_s$ and $-\delta_s \leq \eta_s$.

%Let $u_s =  w_s - e_s$, 
%where $e_s$ is a one-hot vector with value of $1$ at index $s$ and $0$
%elsewhere. 
%We then combine all the coefficient vectors $u_s$ into a matrix:
%$U = \begin{bmatrix} u_{s_1} \cdots u_{s_d} \end{bmatrix}^T$.
%In addition, define $\gamma_s = \tau_s - \ty_s + w_s^T \ty_{-s}$,
%and then $\Gamma = \begin{bmatrix} \gamma_{s_1} \cdots
%  \gamma_{s_d} \end{bmatrix}^T$.
%Similarly, let $\gamma_{s}' = \tau_s + \ty_s - w_s^T \ty_{-s}$, which
%we can group into a matrix $\Gamma'$.
%Constraints~\eqref{c:stealth} can then be written compactly as a
%combination of $U \delta \leq \Gamma$ and $- U \delta \leq \Gamma'$.
%\end{equation*}

Putting everything together, we obtain the following mixed-integer linear programming (MILP)
problem (actually, a minimum over a finite number of these), which solves problem~\eqref{eq:adv} for the linear
regression-based anomaly detector:
\begin{subequations}\label{eq:advlinear}
	\begin{align}
		\nonumber \argmin_{\bar{s}\in S_c} \; &  \min_{\ty, \delta, \alpha} \delta_{\bar{s}} \\
%		\textup{s.t.}\quad & U \delta \leq \Gamma, \; - U
%		\delta \leq \Gamma'  \\
                \textup{s.t.} \quad & \tilde{y}_s - w_s^T
                                      \tilde{y}_{-s} - b_s  \leq
                                      \tau_s, \forall{s} \\
&-\tilde{y}_s + w_s^T
\tilde{y}_{-s} + b_s  \leq \tau_s, \forall{s} \\
&\ty_s = y_s + \delta_s, \forall{s} \\
		& \delta_s \leq M \alpha_s, \forall{s} \\
		& \delta_s \leq \eta_s ;\ -\delta_s \leq \eta_s, \forall{s} \\
		& \sum_{s} \alpha_s \leq B \\
		& \nonumber \ty_s,\delta_s\in \mathbb{R}, \alpha_s \in \{0,1\} \ \forall{s}.
	\end{align}
\end{subequations}
%This optimization problem can be solved using branch-and-bound and KKT conditions. %and local optima can be found 

\subsection{Attacking Neural Network Regression}
% \todo{neural net}

Due to the non-linear structure of a neural network model, Constraints~\eqref{c:stealth} now become non-linear and
non-convex.
To tackle this challenge~\eqref{eq:adv}, we propose an iterative
algorithm, described in Algorithm~\ref{alg:advnn}. 
The algorithm solves the problem by taking small steps in a direction
of increasing objective function.
In each iteration, the algorithm linearizes all the neural networks at
their operating points, and replaces them with the obtained linearized
instances. Then, for each small step, it solves the MILP \eqref{eq:advlinear} to
obtain an optimal stealthy attack in the region around the operating point.
%~\eqref{eq:advlinear} to solve the adversarial regression problem. 
In each iteration, in order to ensure that the obtained solution is
feasible, the algorithm tests the solution with respect to the actual
neural network-based anomaly detectors. 
If it is infeasible, the iteration is repeated using a smaller step
size. 
Otherwise, the same process is repeated until a local optimum is found or we reach a desired maximum number of iterations.

%However, to make sure that the linearized model is a correct representation of the neural network, at each iteration, we only allow the algorithm to only take very small steps from the operation point. 

\begin{algorithm}[t!]
	\caption{Adversarial Regression for Neural Network}
	\label{alg:advnn}
	\footnotesize
	\begin{algorithmic}[1]
		\Statex \textbf{Input}: Measurements $\tilde{y}$, critical sensors $S_c$, budget $B$, algorithm parameters $\epsilon_0$, $n_{max}$%, hyperparameters $\lambda$, $\epsilon$, and $n_{max}$
		%\Statex \textbf{Output}: Perturbations $\delta$ that maximize damage and evade the detectors
		%\State Compute partial derivative of $g(y)$ with respect to each controllable feature
		%\State Store the indices of top $B$ partial derivatives with largest absolute values in set $I$
		%\State $\delta \gets 0$
		\State $\ty_0 \gets \ty$, $\delta_{\bar{s}^\ast} \gets 0$
		\ForAll {$\bar{s} \in S_c$}
		\While {number of iterations $< n_{max}$ \textbf{and}
                  $\epsilon > \epsilon_{min}$}
		\Statex ~~~~~~~~~\texttt{// Linearize all neural networks at $\ty$} \label{goto:linearize}
		%\For{each sensor $s$}
		%\ForAll{$s' \ne s$}
		%\State $w_{ss'} \gets \frac{\partial \fs(\ty_{-s})}{\partial \ty_{s'}}$
		%\EndFor
		%\EndFor
		%\State $W \gets [w_{s_11} \ldots w_{s_1n} ; \ldots ;
                %w_{s_d1} \ldots w_{s_dn}  ]$ \label{goto:W}
                \State $[W,b]
                \gets$\Call{TaylorApproximation}{$f_s(\cdot), \tilde{y}$}
		\Statex ~~~~~~~~~\texttt{// Solve MILP and check feasibility}
		\State $\epsilon \gets \epsilon_0$ 
		\State $\ty' \gets $\Call{Solve\_MILP}{$W$,$b$,$\bar{s}$,$\ty$,$\epsilon$} \label{goto:milp}
		\For{each sensor $s$}
		\If {$| \ty'_s - f^{(s)}(\ty'_{-s}) | > \tau_s$} 
		\State $\epsilon \gets \frac{\epsilon}{2}$	
		\State \Goto{Line~\ref{goto:milp}} 
		\EndIf
		\EndFor
		\State $\ty \gets \ty'$
		\State $\delta_{\bar{s}} \gets \ty'_{\bar{s}}$
		\EndWhile
		\If {$\delta_{\bar{s}^\ast} < \delta_{\bar{s}}$ }
		\State $\bar{s}^\ast \gets \bar{s}$, $\delta_{\bar{s}^\ast} \gets \delta_{\bar{s}}$, $\ty^\ast \gets \ty$
		\EndIf
		\EndFor
		\State \textbf{return } $\ty^\ast$
	\end{algorithmic}
\end{algorithm}

%\todo{linearization and new contraint}
%We now describe Algorithm~\ref{alg:advnn} in more detail. 
%We now describe the algorithm in more detail. 
The algorithm begins with the initial uncorrupted measurement vector
$\ty = y$. For each neural network $\fs$, it obtains a linearized
model using a first-order Taylor expansion around the solution estimate
$\ty$ in the current iteration, which yields a matrix of weights $W$
for the corresponding linear approximation.
%by computing the partial derivative of the output $\fs(\ty)$ with
%respect to the inputs at the current operating point $\ty$, i.e.,
%$\bar{w}^{(s)}_i = \frac{\partial \fs(\ty)}{\partial \ty_i}$ is
%computed for all $i \in S$. 
%Then, for detector $s \in D$, the linearized model can be written as $\bar{f}^{(s)}(\ty + \Delta) = \fs(\ty) + \bar{w}^{(s)^T} \Delta$, where $\Delta \in \mathbb{R}^m$ is a small error vector. 
%$\bar{w}^{(s)} = \{\bar{w}^{(s)}_i \}_{i=1}^{n}$ 
%We denote the coefficients of the linearized models in matrix form by $W = [w^{(s_1)} \ldots w^{(s_d)} ]^T$. 
%\todo{iteration logic}, ADD: adding no dividing by half due to no improvement
Given matrix $W$, we solve the problem by converting it to the
MILP~\eqref{eq:advlinear}, with a small modification:
%The difference is that there is now 
we introduce a constraint that enforces that we only make small
modifications relative to the approximate solution from the previous
iteration.
More precisely, we let $\epsilon \in \mathbb{R}_+$ be the parameter
representing the maximum step size. 
Then, we add the constraint $|\Delta| < \epsilon$ to the MILP, where
$\Delta = y + \delta - \tilde{y}$ and $\tilde{y}$ the latest solution.
%In addition, we update the budget constraint as $\norm{\delta + \Delta}_0 \leq B$, where $\Delta$ is the error vector added at the current iteration, and $\delta = \ty - y$ is the error added in the previous iterations.

Let $\ty'$ be the solution obtained by solving the MILP. We check
whether this solution is feasible by performing forward-propagation in
all the neural networks and checking that no stealth constraint
is violated. If a stealth constraint is violated, which means
that our linearized model was not a good approximation of the neural
network, we discard the candidate solution, reduce the maximum step
size parameter $\epsilon$ to improve the approximation, and re-solve
the MILP. We repeat this process until a feasible solution is found,
in which case the same steps are repeated for a new operating point,
or until we reach the maximum number of iterations $n_{max}$. Finally,
we check whether the solution that is found for the current target
sensor $\bar{s}$ outperforms the solution for the best target sensor
found so far. The algorithm terminates after considering all the
target sensors and returns the best attack vector found.

\subsection{Attacking the Ensemble Model}
%\todo{in this section and why multiple detector}

%\todo{what does this mean for adv regression}
%\paragraph*{Adversarial Regression} 
We can view the ensemble as a single neural network that connects a
perceptron (i.e., the linear regression model) with our initial neural
network at the output layer. Thus, to solve the adversarial regression
problem, we can use the same approach as in the case of neural
networks by modifying Algorithm~\ref{alg:advnn} to obtain linear
approximation parameters $W = \frac{1}{2}(W_{NN} + W_{LR})$ and $b =
\frac{1}{2}(b_{NN} + b_{LR})$, where $W_{LR}$ and $b_{LR}$ are the
parameters of the linear regression model in the ensemble, and
$W_{NN}$ and $b_{NN}$ the parameters of the Taylor approximation of the
neural network piece of the ensemble.
%That is, 
%in
%line~\ref{goto:W}, 
%we compute the linearized coefficient matrix as $W = \frac{1}{2}(W_{NN} + W_{LR})$ where $W_{NN}$ is the matrix of linearized weights of the neural networks obtained by computing the partial derivatives with respect to input, and $W_{LR}$ is the fixed weights of the linear regression model. 

%Note another detector can be implemented considering that checks if $|\hat{y}_{LR} - \hat{y}_{NN}| < d$.
%\begin{tabular}{l} \textbf{if} $|\hat{y}_{LR} - \hat{y}_{NN}| < d$\\ \hspace{.3cm}$\hat{y}_E = \frac{1}{2}(\hat{y}_{LR} + \hat{y}_{NN})$ \\ \textbf{else} \\ \hspace{.3cm} detection alarm\\ \end{tabular}

\section{Resilient Detector}\label{sec:resilience}

Having developed stealthy attacks on the anomaly detection system we
described earlier, we now consider the problem of designing robustness
into such detectors.
Specifically, we allow the defender to tune the thresholds $\tau_s$ to
optimize robustness, accounting for stealthy attacks.
More precisely, we model the game between the defender and attacker as
a \emph{Stackelberg game} in which defender first commits to a
collection of thresholds $\tau_s$ for the anomaly detectors, and the
attacker then computes an attack following the model in
Section~\ref{sec:advregression}.
The defender aims to minimize the impact of attacks
$D(\tilde{y}(\tau))$, subject to a constraint that the number of
false alarms is within $\gamma$ of that for a baseline $\bar{\tau}$,
typically set to achieve a target number of false alarms without
consideration of attacks.
The constraint reflects the common practical consideration that we
wish to detect attacks without increasing the number of false alarms
far beyond a target tolerance level.
We seek a Stackelberg equilibrium of the resulting game, which is
captured by the following optimization problem for the defender:
\begin{align}
\label{E:optdef}
\begin{split}
\min_\tau \max_{s \in S_c} &\quad D_s(\tilde{y}(\tau))\\
\mathrm{s.t.:}& \ \ FA(\tau) \le FA(\bar{\tau}) + \gamma, \tilde{y}(\tau)\ 
  \mathrm{solves}\ \mathrm{Problem}~\eqref{eq:adv},
\end{split}
\end{align}
where $FA(\tau)$ is the number of false alarms when the threshold
vector is $\tau$.

Given thresholds $\tau$, we define the \textit{impact of attack} on a
sensor $s$ as
the mean value of problem~\eqref{eq:adv} over a predefined time period
$T$: $D_s(\tilde{y}) = \sum_{t=1}^T (\ty_s^t - y_s^t)$.
%$D_s(\tilde{y}) = \sum_{t=1}^T \ty_{\bar{s}^\ast}^t - y_{\bar{s}^\ast}^t$.,
%where $\bar{s}^\ast$ is the critical sensor subject to the worst attack.
%To evaluate false alarms, we let $T$ be the duration of an experiment
%under normal operation. 
Similarly, we compute $FA(\tau)$ as the total number of false alarms
over $T$ time steps: $FA(\tau) = \sum_s \sum_{t=1}^{T} 1_{| y_s^t - \fs(y_{-s}^t) | > \taus }$.

The key way to reduce the attack impact $D(\tilde{y}(\tau))$ is by decreasing
detection threshold. This can effectively force the attacker to launch
attacks that follow the behavior of the physical system more closely,
limiting the impact of attacks.
On the other hand, smaller values of the threshold parameters may
result in a higher number of false alarms.
Next, we present a novel algorithm for resilient detector threshold
selection as defined in Problem~\eqref{E:optdef}.
%in the context of stealthy sensor attacks described above
%which balances these two goals as formalized in Problem~\eqref{E:optdef}.

Let $\tau$ be initialized based on a target number of false alarms.
In each iteration, given threshold values $\tau$ computed in the
previous iteration, we find the (approximately) optimal stealthy
attack $\tilde{y}(\tau)$, with associated impact $D(\tilde{y}(\tau))$.
Let $A^\ast$ be the set of critical sensors upon which the attack
$\tilde{y}$ has the largest impact, 
and let $A_{min}$ be the set of sensors with the smallest stealthy attack impact.
To reduce $D_{A^\ast}(\tilde{y}(\tau))$, we reduce $\tau_{\bar{s}}$ for each $\bar{s} \in
A^\ast$.
Then, to minimize the change in the number of false alarms, we increase the threshold $\tau_s$ for each $s
\in A_{min}$ to compensate for the false alarms added by detectors of
sensors in $A^\ast$. 
These steps are repeated until we reach a local optimum.
We formalize this in Algorithm~\ref{alg:threshold}.

\begin{algorithm}[t!]
	\caption{Resilient Detector}
	\label{alg:threshold}
	\footnotesize
	\begin{algorithmic}[1]
		%\Statex \textbf{Input}: budget $B$ %Measurements and features $\tilde{y}$, target sensor $a$%, hyperparameters $\lambda$, $\epsilon$, and $n_{max}$
		\Statex \textbf{Initialize}: Initial threshold setting
                $\bar{\tau}$, and initial $\epsilon$
                \State $\tau' = \tau = \bar{\tau}$
                \State $\ty \gets \Call{Attack}{\tau}$
                \State $I = \infty$
		\While {number of iterations $< n_{max}$}
                \If {$\max_{s \in S_c} D_s(\ty) \le I$}
                \State $\tau \gets \tau'$
                \Else 
                \State $\epsilon \gets \frac{\epsilon}{2}$
                \EndIf
                \State $I \gets \max_{s \in S_c} D_s(\ty)$
		\State $A^\ast \gets \argmax_{s \in S_c} D_s(\ty)$
		\State $A_{min} \gets \argmin_{s \in S_c} D_s(\ty)$
		\ForAll {$s \in A^\ast$}
		\State $\tau_{s}' \gets \tau_{s} - \epsilon$
		\EndFor
		\State $\Delta FP \gets FP(\tau') - FP(\tau)$
		\ForAll {$s \in A_{min}$}
		\State $\tau_s' \gets FP_{s}^{-1}(FP_{s}(\tau_s)- \frac{1}{|A_{min}|}\Delta FP)$
		\EndFor
		\State $\ty \gets \Call{Attack}{\tau'}$
		\EndWhile
		\State \textbf{return } $\tau$
	\end{algorithmic}
\end{algorithm}
 
\section{Experiments}\label{sec:experiment}

We evaluate our contributions using a case study of the well-known Tennessee-Eastman process control system (TE-PCS)
%a safety-critical process control systems.  
%In particular, we study the well-known Tennessee-Eastman process control system (TE-PCS). 
First, we design regression-based anomaly detectors for the critical sensors in the system (e.g., pressure of the reactor, level of the product stripper). Then, we consider scenarios where an adversary launches sensor attacks in order to drive the system to an unsafe state. Finally, we evaluate the resilience of the system against such attacks using baseline detectors and the proposed resilient detector.

\subsection{Tennessee-Eastman Process Control System}

% te-pcs architecture and products
We present a brief description of the Tennessee-Eastman Process Control System (TE-PCS). TE-PCS involves two simultaneous gas-liquid exothermic reactions for producing two liquid products~\cite{downs1993plant}. The process has five major units: reactor, condenser, vapor-liquid separator, recycle compressor, and product stripper. The chemical process consists of an irreversible reaction which occurs in the vapor phase inside the reactor. Two non-condensible reactants A and C react in the presence of an inert B to form a non-volatile liquid product D. The feed stream 1 contains A, C, and trace of B; feed stream 2 is pure A; stream 3 is the purge containing vapors of A, B, and C; and stream 4 is the exit for liquid product~D. 
%See Figure~\ref{fig:te} in Appendix~\ref{sec:appendix_te} for a diagram of the TE-PCS~\cite{bathelt2015revision}.

%\noindent{\bf Safety Constraints and Control Objectives}
There are 6 safety constraints that must not be violated (e.g, safety limits for the pressure and temperature of reactor). These safety constraints correspond to 5 critical sensors: pressure, level, and temperature of the reactor, level of the product stripper, and level of the separator. Further, there are several control objectives that should be satisfied, e.g., maintaining process variables at desired values and keeping system state within safe operating conditions. The monitoring and control objectives are obtained using 41 measurement outputs and 12 manipulated variables.

%\noindent{\bf Simulation of Sensor Attacks}
We use the revised Simulink model of TE-PCS~\cite{bathelt2015revision}. We consider the implementation of the decentralized control law as proposed by~\citeauthor{ricker1996decentralized}~\shortcite{ricker1996decentralized}. To launch sensor attacks against TE-PCS, we update the Simulink model to obtain an information flow 
%similar to Figure~\ref{fig:diagram}.
%as shown in Figure~\ref{fig:tesimulink} Appendix~\ref{sec:appendix_te}. 
%That is, 
in which the adversary receives all the sensor measurements and control inputs, solves the adversarial regression problem, and then adds the error vector to the actual measurements. 
%Note that the detector receives the control inputs and the attacked measurements.

\begin{figure}
\centering
\includegraphics[width=0.65\linewidth]{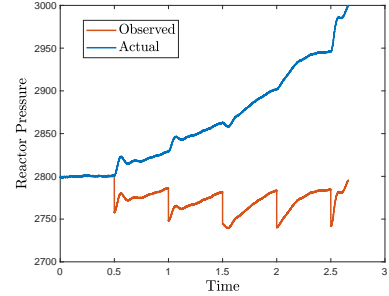}
\caption{Pressure of the reactor when a sensor attack starts at $t = 0.5$. After $2$ hours the pressure reaches an unsafe state.} %$B=15$ linear regressor
\label{fig:fig-attack_r_t2b15p5}
\end{figure}

% unsafety due to attack
Figure~\ref{fig:fig-attack_r_t2b15p5} shows how a sensor attack may drive the system to an unsafe state. In this scenario, the pressure of the reactor exceeds 3000 kPa which is beyond the safety limit and can result in reactor explosion. 

\subsection{Regression-Based Anomaly Detector}

%\paragraph*{Collection of Normal Data}  
% data collection, sampling, initial state
To protect the system against sensor attacks, we build a detector for each critical sensor.
To train the detectors
%predictors for the machine learning regression-based detectors, we need data that correctly represent system behavior in different operation modes. To do so, 
we run simulations that model the system operation for 72 hours, and collect the sensor measurements and control inputs. Each simulation consists of 7200 time steps and thus, for each simulation scenario, we record 7200$\times$53 datapoints. To make sure that the dataset represents different modes of operation, we repeat the same steps for different initial state values. We consider a total of 20 different initial state values which gives us 20$\times$7200$\times$53~$\approx$~7.5 million datapoints. 
%Three of the control inputs are always constant and so effectively we only store 50 values at each timestep. 

%We train the predictors as described below. Then, given the predictors, we implement the detector using~\eqref{eq:threshold_test}.

\begin{figure*}[t!]
\centering
\begin{minipage}[c]{0.9\linewidth}
\centering
\subfloat[]{\includegraphics[width=0.325\linewidth]{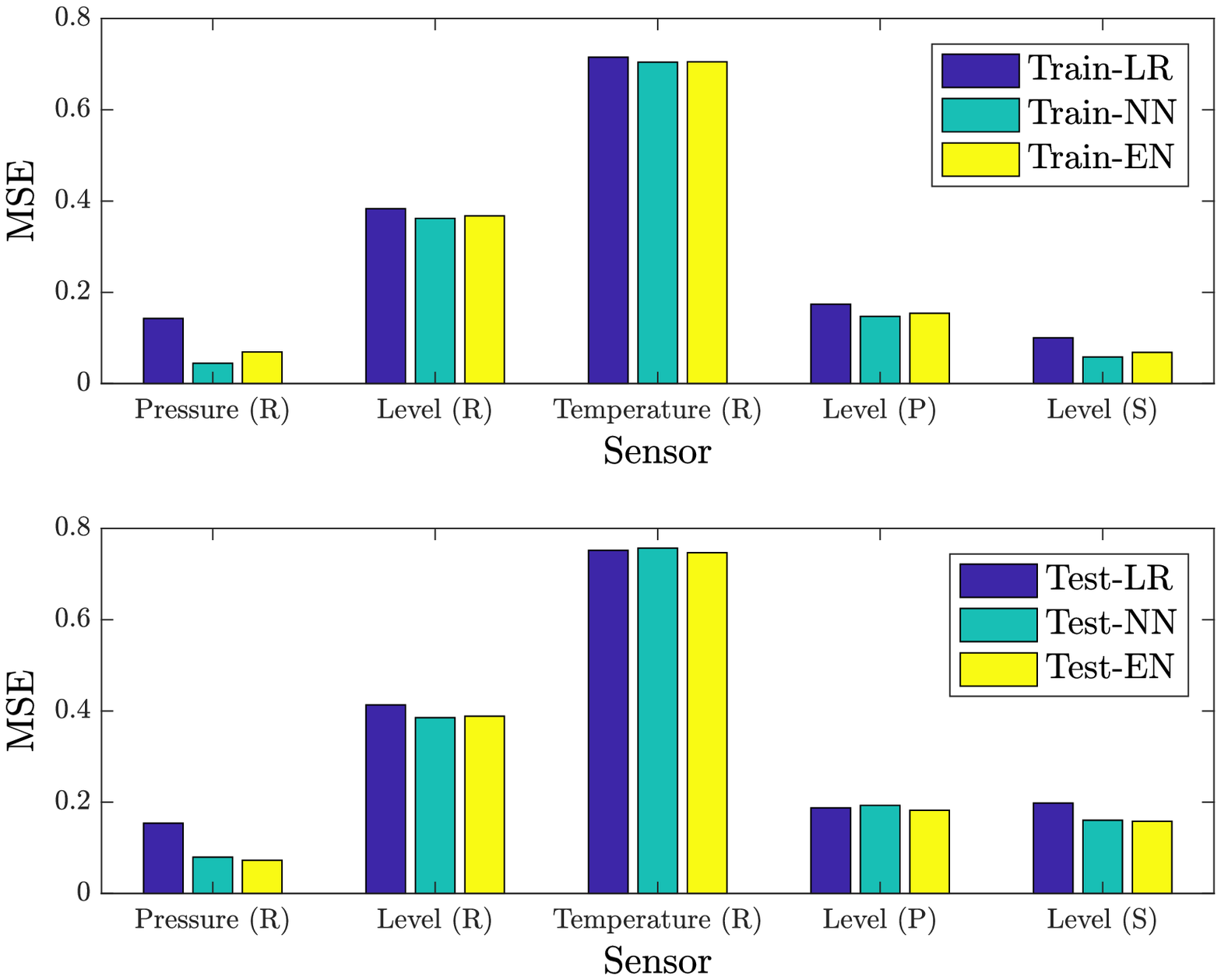} \label{fig:mse} \label{fig:mse_lr} \label{fig:mse_en} \label{fig:mse_nn}}
\hfil
\subfloat[]{\includegraphics[width=0.325\linewidth]{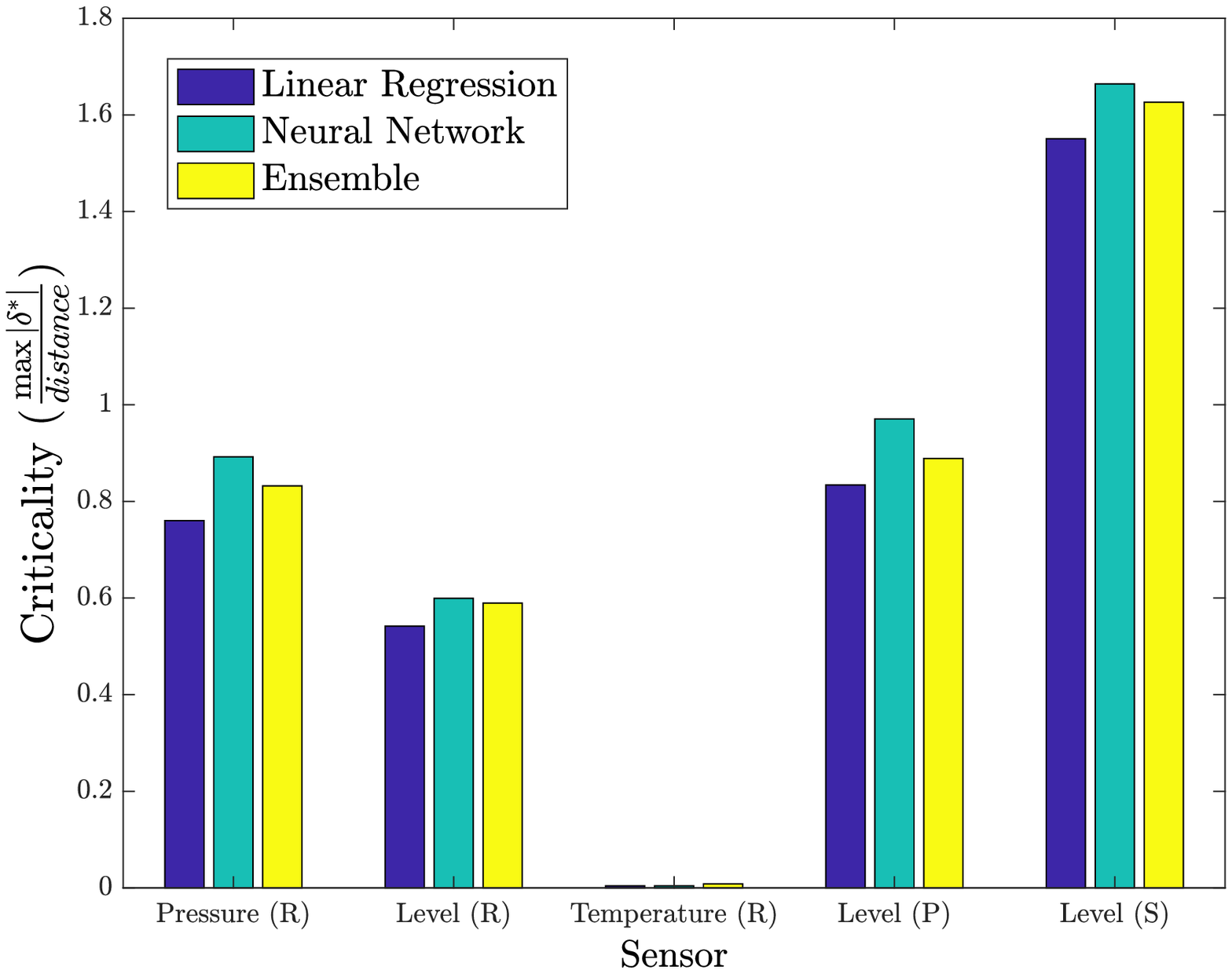} \label{fig:critical}}
\hfil
\subfloat[]{\includegraphics[width=0.325\linewidth]{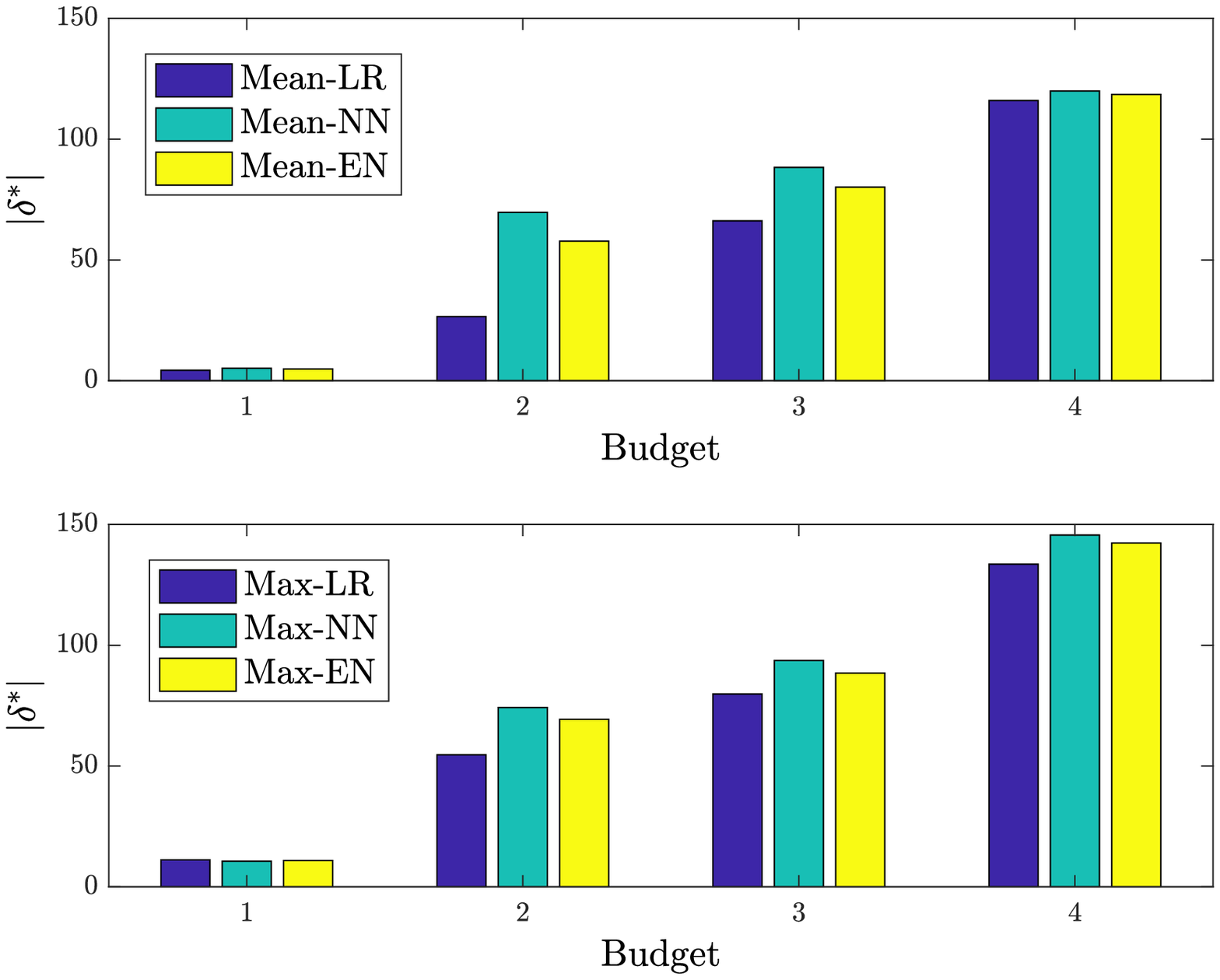} \label{fig:budget}}
\end{minipage}
\caption{(a) MSE of Linear Regression, Neural Network, and Ensemble Model. All metrics are computed using Normalized Data. (b) Stealthy attack solution for the five critical sensors and different detectors. (c) Stealthy attack solution for the pressure of the reactor considering varying budgets.}
\end{figure*}

\paragraph{Linear Regression-Based Detector}
% lr
Using our collected data, we train linear regression models for the critical sensors. We use the current value of the remaining 36 non-critical sensors as well as the 9 non-constant control inputs as features of the model. Figure~\ref{fig:mse_lr} shows the performance of the linear regression predictor on training and test data. Note that since the data is sequential, the train and test data cannot be randomly sampled and instead, we divide the data in two blocks. Also, to be able to compare the performance of predictors trained for different variables, we compute the MSE for normalized values instead of the actual values.

\begin{figure*}[t!]
\centering
\begin{minipage}[c]{0.9\linewidth}
\centering
\subfloat[]{\includegraphics[width=0.325\linewidth]{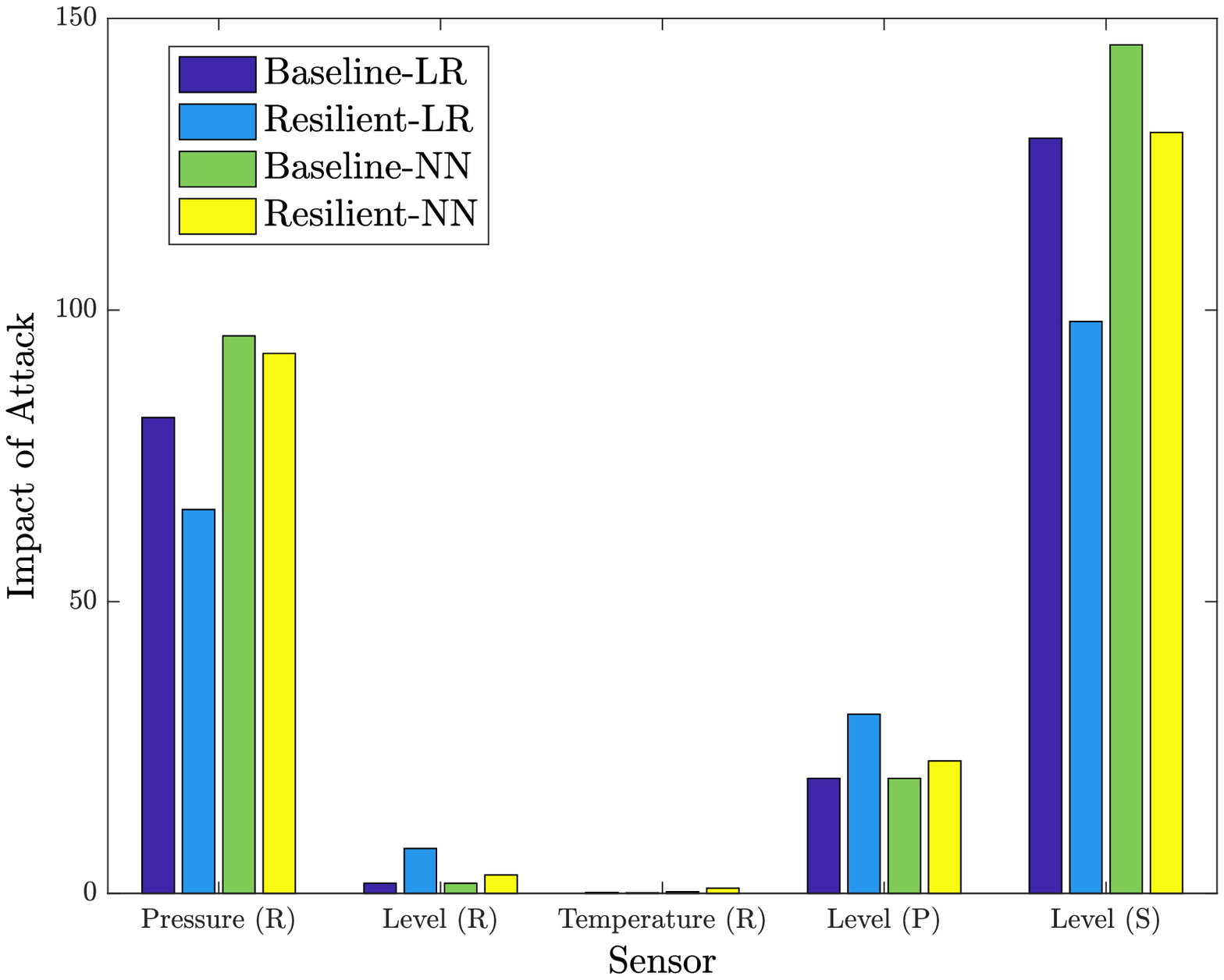} \label{fig:res_impact}}
\hfil
\subfloat[]{\includegraphics[width=0.325\linewidth]{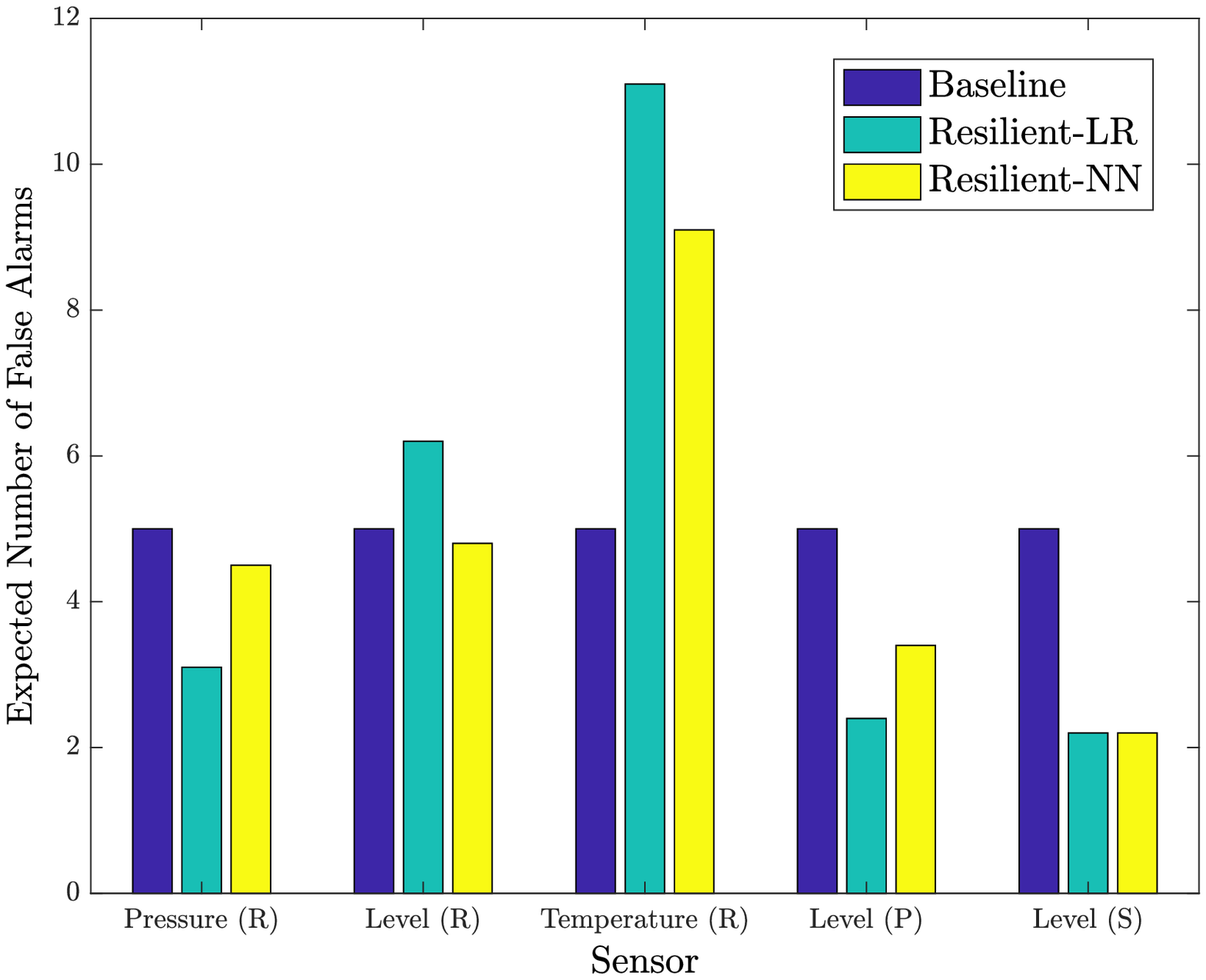} \label{fig:res_fp}}
%\hfil
%\subfloat[]{\includegraphics[width=0.32\linewidth]{figure/fig-tradeoff} \label{fig:tradeoff}}
\end{minipage}
\caption{Resilient Detector for Linear Regression. (a) Impact of Resilient Detector compared to baseline. (b) False positive of Resilient Detector compared to baseline. (c) Trade-off between stealthy attack impact and false positive rate.}
\label{fig:resilient}
\end{figure*}

%\begin{table}
%\caption{MSE of Test Data (Original Scale)} \smallskip 
%\label{tab:estimation} 
%\centering
%\begin{tabular}{ l  c  c  c  c  c}
%	\toprule
%	& PR & LR & TR & LP & LS \\
%	\midrule
%	LR-MSE & $9.05$ & $0.26$ & $0.00$ & $1.53$ & $7.17$\\
%	NN-MSE & $6.28$ & $0.25$ & $0.00$ & $1.41$ & $5.90$\\
%	EN-MSE & $5.74$ & $0.26$ & $0.00$ & $1.33$ & $5.81$\\
%	\bottomrule
%\end{tabular}
%\end{table}

\paragraph{Neural Network-Based Detector}
%network architecture, dropout, L2 regularization, batch normalization
Next, we train the neural network regression models for the critical sensors. Unlike linear regression models, neural networks require several hyperparameters to be selected (e.g., network architecture, activation function, optimization algorithm, regularization technique). 
We considered neural networks with 2 to 4 hidden layers and 10 to 20 neurons in each layer. All the neurons in the hidden layers use tanh activation functions. We also experimented with ReLU activation functions but tanh performs better.
We trained the networks in Tensorflow for 5000 epochs using Adam optimizer with $\beta_1 = 0.9$, $\beta_2 = 0.999$, and $\epsilon = 10^{-8}$, and a learning rate of $0.01$. Figure~\ref{fig:mse_nn} shows the MSE for training and test data, which outperforms the linear regression model.
%Note that we may be able to obtain better results through more fine-tuning, but this result is satisfactory.
%We trained five neural networks, one for each critical sensor, in TensorFlow using our dataset and the same set of features. 
%\paragraph*{Ensemble} 
% ADD: result?
Finally, Figure~\ref{fig:mse_nn} shows the result for the ensemble model. %we combine the results of linear regression and neural network using a bagging approach as depicted in Figure~\ref{fig:ensemble}.

\subsection{Attacking Anomaly Detection}

%We solve the considering linear regression-, neural network-, and the ensemble-based detectors. 
%\paragraph*{Comparison of Critical Sensors}
Figure~\ref{fig:critical} shows the attack solution for each critical sensor considering different detectors. As it can be seen, the temperature of the reactor is the least critical sensor while the level of the stripper is the most critical. %OR critical sensor
%X: critical sensor. Y: max deviation/average deviation. Legend: LR, NN, EN
Interestingly, neural network models, as well as ensembles, tend to be more vulnerable---often significantly---then the linear regression-based detectors.
This is reminiscent of the observed vulnerability of neural network models to adversarial tampering~\cite{goodfellow2014explaining}, ours is the first observation that neural networks can be \emph{more} fragile than simple linear models.
%The only interesting exception is the product stripper sensor, which is indeed substantially more vulnerable to attacks.

\subsection{Resilient Detector}
%\paragraph*{Trade-off Between Attack Impact and False Alarms}
%Figure~\ref{fig:tradeoff} shows the trade-off between false positive and attack impact while considering same threshold for all detectors.

%\paragraph*{Threshold Selection}
We use the resilient detector algorithm to find threshold values that reduce the stealthy attack impact as described in Section~\ref{sec:resilience}.
We do this in the context of linear regression.
Let $T^\ast = 1$ hour, be the desired value for the expected time between false alarms for all detectors. As a baseline, for each detector $s$, we set threshold values $\tau_s = FP_s^{-1}(\frac{T^\ast}{d})$. %This result in attack impacts as shown in Figure~\ref{fig:resilient}. 

Then, we use Algorithm~\ref{alg:threshold} to change thresholds in order to obtain better resilience. In the first iteration, stripper level is the most critical sensor, and reactor temperature the least critical.
Therefore, we decrease the threshold corresponding to stripper level and increase the value of threshold for reactor temperature. We repeat these steps using Algorithm~\ref{alg:threshold} until we obtain results shown in Figure~\ref{fig:resilient}. As we can see, we can significantly reduce the impact of the stealthy attack compared to the baseline (Figure~\ref{fig:res_impact}) without increasing the total number of false alarms (Figure~\ref{fig:res_fp}, where the total number of false alarms is the sum over all sensors).

%2) X:time. Y: deviation. Legend: B

\section{Conclusions}\label{sec:conclusion}
We studied the design of resilient anomaly detection systems in
CPS. We considered a scenario where the CPS is monitored by machine
learning regression-based anomaly detectors, and an omniscient
adversary capable of perturbing the values of a subset of sensors. The
adversary's objective is to lead the system to an unsafe state (e.g.,
raising the pressure of a reactor in a process control system beyond
its safety limit) without being detected. We compute an optimal
stealthy attack for linear regression, neural network, and a simple
ensemble of these. Finally, we present an approach to mitigate the
impact of stealthy attacks through resilient configuration of detector
thresholds. We demonstrated the effectiveness of our methods using a
case study of a process control system.

\section*{Acknowledgments}

%% ARO, ONR, NSF (CNS, CAREER), NIH RISK
This research was partially supported by the National  Science
Foundation  (CNS-1238956, CNS-1640624, IIS-1649972, OISE-1743772), Air Force
Research Laboratory (8750-14-2-0180),  Office of  Naval
Research  (N00014-15-1-2621),  Army  Research  Office
(W911NF-16-1-0069), National Institutes of Health (R01HG006844-05), and National Institute of Standards and Technology
(70NANB17H266).

%\nocite{*}
%\bibliographystyle{ACM-Reference-Format}  % do not change this line!
%\bibliographystyle{named}
%\bibliography{reference}  % put name of your .bib file here

\begin{thebibliography}{}

\bibitem[\protect\citeauthoryear{Bathelt \bgroup \em et al.\egroup
  }{2015}]{bathelt2015revision}
Andreas Bathelt, N~Lawrence Ricker, and Mohieddine Jelali.
\newblock Revision of the tennessee eastman process model.
\newblock {\em IFAC-PapersOnLine}, 48(8):309--314, 2015.

\bibitem[\protect\citeauthoryear{Biggio \bgroup \em et al.\egroup
  }{2010}]{biggio2010multiple}
Battista Biggio, Giorgio Fumera, and Fabio Roli.
\newblock Multiple classifier systems for robust classifier design in
  adversarial environments.
\newblock {\em International Journal of Machine Learning and Cybernetics},
  1(1-4):27--41, 2010.

\bibitem[\protect\citeauthoryear{Biggio \bgroup \em et al.\egroup
  }{2014}]{Biggio13}
Battista Biggio, Giorgio Fumera, and Fabio Roli.
\newblock Security evaluation of pattern classifiers under attack.
\newblock {\em IEEE Transactions on Knowledge and Data Engineering},
  26(4):984--996, 2014.

\bibitem[\protect\citeauthoryear{C{\'a}rdenas \bgroup \em et al.\egroup
  }{2011}]{cardenas2011attacks}
Alvaro~A C{\'a}rdenas, Saurabh Amin, Zong-Syun Lin, Yu-Lun Huang, Chi-Yen
  Huang, and Shankar Sastry.
\newblock Attacks against process control systems: risk assessment, detection,
  and response.
\newblock In {\em Proceedings of the 6th ACM symposium on information, computer
  and communications security}, pages 355--366. ACM, 2011.

\bibitem[\protect\citeauthoryear{Dalvi \bgroup \em et al.\egroup
  }{2004}]{dalvi2004adversarial}
Nilesh Dalvi, Pedro Domingos, Sumit Sanghai, Deepak Verma, et~al.
\newblock Adversarial classification.
\newblock In {\em Proceedings of the tenth ACM SIGKDD International Conference
  on Knowledge discovery and data mining}, pages 99--108. ACM, 2004.

\bibitem[\protect\citeauthoryear{Downs and Vogel}{1993}]{downs1993plant}
James~J Downs and Ernest~F Vogel.
\newblock A plant-wide industrial process control problem.
\newblock {\em Computers \& chemical engineering}, 17(3):245--255, 1993.

\bibitem[\protect\citeauthoryear{Ghafouri \bgroup \em et al.\egroup
  }{2016}]{ghafouri2016optimal}
Amin Ghafouri, Waseem Abbas, Aron Laszka, Yevgeniy Vorobeychik, and Xenofon
  Koutsoukos.
\newblock Optimal thresholds for anomaly-based intrusion detection in dynamical
  environments.
\newblock In {\em International Conference on Decision and Game Theory for
  Security}, pages 415--434. Springer, 2016.

\bibitem[\protect\citeauthoryear{Ghafouri \bgroup \em et al.\egroup
  }{2017}]{ghafouri2017optimal}
Amin Ghafouri, Aron Laszka, Abhishek Dubey, and Xenofon Koutsoukos.
\newblock Optimal detection of faulty traffic sensors used in route planning.
\newblock In {\em International Workshop on Science of Smart City Operations
  and Platforms Engineering (SCOPE)}, pages 1--6, 2017.

\bibitem[\protect\citeauthoryear{Goodfellow \bgroup \em et al.\egroup
  }{2015}]{goodfellow2014explaining}
Ian~J Goodfellow, Jonathon Shlens, and Christian Szegedy.
\newblock Explaining and harnessing adversarial examples.
\newblock In {\em International Conference on Learning Representations}, 2015.

\bibitem[\protect\citeauthoryear{Grosshans \bgroup \em et al.\egroup
  }{2013}]{Grosshans13}
Michael Grosshans, Christoph Sawade, Michael Br\"{u}ckner, and Tobias Scheffer.
\newblock Bayesian games for adversarial regression problems.
\newblock In {\em International Conference on International Conference on
  Machine Learning}, pages 55--63, 2013.

\bibitem[\protect\citeauthoryear{Junejo and Goh}{2016}]{junejo2016behaviour}
Khurum~Nazir Junejo and Jonathan Goh.
\newblock Behaviour-based attack detection and classification in cyber physical
  systems using machine learning.
\newblock In {\em Proceedings of the 2nd ACM International Workshop on
  Cyber-Physical System Security}, pages 34--43. ACM, 2016.

\bibitem[\protect\citeauthoryear{Li and Vorobeychik}{2014}]{li14}
Bo~Li and Yevgeniy Vorobeychik.
\newblock Feature cross-substitution in adversarial classification.
\newblock In {\em Neural Information Processing Systems}, pages 2087--2095,
  2014.

\bibitem[\protect\citeauthoryear{Li and Vorobeychik}{2018}]{Li18}
Bo~Li and Yevgeniy Vorobeychik.
\newblock Evasion-robust classification on binary domains.
\newblock {\em ACM Transactions on Knowledge Discovery from Data}, 2018.
\newblock to appear.

\bibitem[\protect\citeauthoryear{Lowd and Meek}{2005}]{lowd2005adversarial}
Daniel Lowd and Christopher Meek.
\newblock Adversarial learning.
\newblock In {\em Proceedings of the eleventh ACM SIGKDD international
  conference on Knowledge discovery in data mining}, pages 641--647. ACM, 2005.

\bibitem[\protect\citeauthoryear{Nader \bgroup \em et al.\egroup
  }{2014}]{nader2014l_p}
Patric Nader, Paul Honeine, and Pierre Beauseroy.
\newblock $\{l_p\}$-norms in one-class classification for intrusion detection
  in scada systems.
\newblock {\em IEEE Transactions on Industrial Informatics}, 10(4):2308--2317,
  2014.

\bibitem[\protect\citeauthoryear{Ricker}{1996}]{ricker1996decentralized}
N~Lawrence Ricker.
\newblock Decentralized control of the tennessee eastman challenge process.
\newblock {\em Journal of Process Control}, 6(4):205--221, 1996.

\bibitem[\protect\citeauthoryear{Urbina \bgroup \em et al.\egroup
  }{2016}]{urbina2016limiting}
David~I Urbina, Jairo~A Giraldo, Alvaro~A Cardenas, Nils~Ole Tippenhauer, Junia
  Valente, Mustafa Faisal, Justin Ruths, Richard Candell, and Henrik Sandberg.
\newblock Limiting the impact of stealthy attacks on industrial control
  systems.
\newblock In {\em Proceedings of the 2016 ACM SIGSAC Conference on Computer and
  Communications Security}, pages 1092--1105. ACM, 2016.

\bibitem[\protect\citeauthoryear{Vorobeychik and Li}{2014}]{Vorobeychik14}
Yevgeniy Vorobeychik and Bo~Li.
\newblock Optimal randomized classification in adversarial settings.
\newblock In {\em International Conference on Autonomous Agents and Multiagent
  Systems}, pages 485--492, 2014.

\end{thebibliography}

\newpage
\appendix
\section{Proof of Proposition~\ref{prop:nphard}}

\begin{deff} [Adversarial Regression Problem (Decision Version)]
Given an anomaly detector, a set of critical sensors $S_c$, an attack budget $B$, and a threshold attack $\ty_c^\ast$, determine whether there exists an attack with a value of at least $\ty_c^\ast$.
\end{deff}

We prove this proposition using a reduction from a well-known NP-hard problem, the Maximum Independent Set Problem.

\begin{deff}[Maximum Independent Set Problem (Decision Version)]
Given an undirected graph $G = (V, E)$ and a threshold cardinality $k$, determine whether there exists an independent set of nodes (i.e., a set of nodes such that there is no edge between any two nodes in the set) of cardinality~$k$.
\end{deff}

\begin{proof}
Given an instance of the Maximum Independent Set Problem (MIS), that is, a graph $G = (V, E)$ and a threshold cardinality $k$, we construct an instance of the Adversarial Regression Problem (ARP) as follows:
1) Let the set of sensors be $S := V \cup \{c\}$, where $c$ is not connected to any node, and let $S_c := \{c\}$ be the only critical sensor in the system. 
2) Let $D := S$. For each detector $s \in D$, let $\tau\s := 0$. Further, let $\fs(\ty_{-s})$ be the number of nonzero elements in $\ty_{-s}$, if $\ty_s \not = 0$ and those elements form an independent set, and zero otherwise. 
3) Let $y_s := 0$ for every sensor $s \in S$.
4) Let $B := k+1$. %$\delta = \{0, 1, \ldots, |S|\}$.
%This function can easily be computed in polynomial time: iterate through the elements, and for each element, test whether an edge exists in the graph.
5) Finally, let the threshold objective be $y_c^* := k + 1$.

Clearly, the above reduction can be performed in polynomial time. Hence, it remains to show that the constructed instance of ARP has a solution \emph{if and only if} the given instance of MIS does.

First, suppose that MIS has a solution, that is, there exists an independent set $A$ of $k$ nodes. We claim that the set $A' = A \cup \{c\}$ where $\ty_s = k+1$ for every $s \in A'$, and $\ty_s = 0$ otherwise, is a solution to ARP. %That is, we can attack the sensors in $A$ and obtain the threshold attack measurement $k+1$.
For nonzero sensors, since $A'$ is independent, the value of~$\fs(\ty_{-s})$ is equal to the number of nonzero sensors in $A'$, which is equal to $k+1$.
%Consequently, for all sensors in $A'$, we can assign $\ty_s = k+1$,
This satisfies the stealthiness constraint since for nonzero sensors $|\ty_s - \fs(\ty_{-s})| = |k+1-(k+1)| = 0\leq \tau\s$, and for zero sensors $|\ty_s - \fs(\ty_{-s})| = |0-0| = 0\leq \tau\s$. Clearly, the budget constraint is also satisfied, and so $y_c^\ast = k+1$ is obtained.

Second, suppose that MIS has no solution, that is, every set of at least $k$ nodes is non-independent.
As a consequence, we have $\fs(\ty_{-s}) < k + 1$ for every detector, since otherwise, there would exist a set of at least $k + 1$ nodes (which could include $c$) that are independent of each other, which would contradict our supposition. Since $|\ty_c - f_c (\ty_{-c})| < \tau_c = 0$ implies that $\ty_c = f_c$, ARP cannot have a solution, which concludes our proof.
\end{proof}

%\appendix
%\input{appendix}

\end{document}